# Evolutionary Optimization in an Algorithmic Setting

**Mark Burgin**

Dept. of Mathematics
Univ. of California
405 Hilgard Avenue
Los Angeles, CA 90095

**Eugene Eberbach**

Comp. and Inf. Science Dept.
Univ. of Massachusetts
285 Old Westport Road
North Dartmouth, MA 02747

**Abstract**

Evolutionary processes proved very useful for solving optimization problems. In this work, we build a formalization of the notion of cooperation and competition of multiple systems working toward a common optimization goal of the population using evolutionary computation techniques. It is justified that evolutionary algorithms are more expressive than conventional recursive algorithms. Three subclasses of evolutionary algorithms are proposed here: bounded finite, unbounded finite and infinite types. Some results on completeness, optimality and search decidability for the above classes are presented. A natural extension of Evolutionary Turing Machine model developed in this paper allows one to mathematically represent and study properties of cooperation and competition in a population of optimized species.

**Keywords:** combinatorial optimization, evolutionary computation, complexity theory, computer science, global optimization, search theory, cost benefit analysis

## 1. Introduction

Combinatorial optimization is aimed at finding optimal solutions of complex search problems. It can be categorized into exact and heuristic methods. Exact methods consist of branch and bound, dynamic programming, Lagrangian relaxation, and linear and integer programming. Heuristic methods contain evolutionary algorithms, tabu search, ant colony optimization, particle swarm optimization, and simulated annealing.

In this paper, we argue that the classification of methods for combinatorial optimization into exact and heuristic classes is a little superficial. Everything depends on the complexity of the search problem. So called, exact methods, can and have to be interrupted to produce approximate solutions for large search problems. For example, this can be appreciated if somebody tries to use dynamic programming to solve some of NP-complete problems. Namely, it is possible to use "exact" dynamic programming for 6-10 cities in traveling salesman problem, but only inexact dynamic programming solutions for hundreds and thousands cities are tractable. Although "inexact" evolutionary algorithms and simulated annealing methods can guarantee to find exact solutions for traveling salesman problem, in a general case, this is guaranteed only in infinite number of generations. However, solving a problem in infinite number of steps goes beyond classical algorithms and Turing Machines, and in spite of being common in mathematics, encounters steady resistance in algorithmic-based conventional computer science.

In this paper, we show how to achieve the same results, i.e., to find exact solutions for hard problem, in finite time (number of steps). Namely, we can use super-recursive algorithms, which allow one to solve many problems undecidable in the realm of recursive algorithms (Burgin, 2005). We argue that it is beneficial for computer science to go beyond recursive algorithms, making possible to look for exact solutions of intractable problems or even to find solutions of undecidable problems, whereas recursive solutions do not exist. As the basic computational model, we take evolutionary Turing machines introduced in (Eberbach, 2005a) but extend their computational power by allowing to use not only Turing machines in a row, but also inductive Turing machines and limit Turing machines, which are more powerful than Turing machines.

In this paper, we study the following problems for the finite types of evolutionary computations: completeness, optimality, search optimality, total optimality and decidability for single and multiple cooperating or competing individuals.

The paper is organized as follows. In section 2 we give a short primer on problem solving. Section 3 introduces inductive and limit Turing machines as substantially more expressive extensions of the Turing machine model. Section 4 presents Evolutionary Algorithms and an evolutionary Turing machine as a formal model of evolutionary computation. In section 5, three subclasses of evolutionary computation are defined: bounded finite, unbounded finite and infinite. Results on completeness, optimization convergence and decidability for these three subclasses are obtained in Section 6. A formal model for cooperating and competitive population agents trying to achieve a common goal is developed in Section 7. Section 8 contains conclusions and problems to be solved in the future.

## 2. Primer on Problem Solving

An algorithm provides a recipe to solve a given problem. For simplicity, we assume that an algorithm consists of a finite number of rules, each having easily comprehensible, well defined and implementable meaning. In addition, we consider such algorithms that have states like Turing machines or finite automata.

All algorithms are divided into three big classes (Burgin, 2005): *subrecursive*, *recursive*, and *super-recursive*.

**Definition 2.1.** Algorithms and automata that have the same computing/accepting power (cf., (Burgin, 2005a)) as Turing machines are called *recursive*.

Examples are partial recursive functions or random access machines.

**Definition 2.2.** Algorithms and automata that are weaker than Turing machines are called *subrecursive*.

Examples are finite automata, context free grammars or push-down automata.

**Definition 2.3.** Algorithms and automata that are more powerful than Turing machines are called super-recursive.

Examples are inductive Turing machines, Turing machines with oracles or finite-dimensional machines over the field of real numbers.

A recursive algorithm starts from the initial state and terminates (if successful) in one terminal state from the set of terminal/goal states. If the algorithm reaches its goal state, then we say that the algorithm satisfied its goal and problem has been solved. The goal test (also called a termination condition) determines whether a given state is a goal state.

The set of states that are reachable from the initial state forms the search space of the algorithm. However, there is no guaranty that any solution of the problem belongs to this space or even to the space of all states. Thus, the solution of the problem can be interpreted as a search process through the set of states. The state space forms a directed graph (or its special case - a tree) in which nodes are states and the arcs between nodes are actions. Search can be deterministic or nondeterministic/probabilistic. Multiple solutions can be ranked using an objective function (e.g., a utility or fitness function). In particular, there can be none, one, or several optimal solutions to the problem. Using objective functions allows capturing the process of iterative approximation of solutions and different qualities of solutions in contrast to simply a binary decision: a goal is reached or not. In some cases, it is impossible to reach the goal, and we need approximations from the very beginning.

The performance of search algorithms can be evaluated in four ways (see, e.g., (Russell and Norvig, 1995) capturing three criteria: whether a solution has been found, its quality and the amount of resources used to find it.

**Definition 2.4.** (*Completeness, optimality, search optimality, and total optimality*) We say that the search algorithm is

- *Complete* if it guarantees reaching a terminal state/solution if there is one.
- *Optimal* if it finds the solution with the optimal value of its objective function.
- *Search Optimal* if it finds the solution with the minimal amount of resources used (e.g., minimal time or space complexity).
- *Totally Optimal* if it finds the solution both with the optimal value of its objective function and with the minimal amount of resources used.

Let us consider an optimization problem *P* and two spaces: *optimizing* and *optimization* space. In an optimization space *X*, elements are representations of those objects that are optimized. For instance, elements from the optimization space *X* are fixed binary strings for genetic algorithms (GAs), Finite State Machines for evolutionary programming (EP), parse trees for genetic programming (GP), vector of real numbers for evolution strategies (ES),

There are also relations between elements from *X*. These relations can represent relations between objects that are optimized or correspond only to elements from *X*.

Usually, it is assumed that that the optimization space *X* contains all representations of some form for all possible optimized elements (species) of a given kind. In practice, only some finite part of the space *X* is considered. However, for being able to treat an optimization problem by mathematical tools, we need, as a rule, to take an infinite space *X*. For instance, in each step of a classical optimization schema (procedure), only a finite number of species and their representations are involved. These species form an optimization pool of the corresponding generation.

In an optimizing space **A**, elements are optimization algorithms. There are also relations between elements from **A**. Thus, **A** is a kind of an algorithm space. For instance, the set **T** of all Turing machines with the binary relation "the machine $T_1$ has more computing power than the machine $T_2$" (cf., (Burgin, 2005a)) is an algorithm space. All genetic algorithms form an optimizing space.

The optimization space *X* contains a solution subspace $X_F$. Elements from $X_F$ are solutions to the problem *P*. The optimizing space **A** contains a solving subspace $\mathbf{A}_F$. Algorithms from $\mathbf{A}_F$ give solutions to the problem *P*.

Let ***R*** be the set of all real numbers and ***R***$^+$ be the set of all non-negative real numbers.

**Definition 2.5.** (*Problem solving as a multiobjective optimization problem*)

Given an objective function *f*: **A** × *X* → ***R***$^+$, problem-solving can be considered as a *multiobjective minimization problem* to find $A^* \in \mathbf{A}_F$ and $x^* \in X_F$ such that

$f(A^*, x^*) = \min\{f_1(f_2(A), f_3(x)), A \in \mathbf{A}, x \in X\}$

where $f_3$ is a *problem-specific objective function*, $f_2$ is a *search algorithm objective function*, and $f_1$ is an *aggregating function* combining $f_2$ and $f_3$.

Without losing generality, it is sufficient to consider only minimization problems. However, traditionally many problems are treated as maximization problems. An objective function $f_3$ can be expanded to multiple objective funtions if problem considered has several objectives. The aggregating function $f_1$ can be arbitrary (e.g., additive, multiplicative, a linear weighted sum). The only requirement is that $f_1$ becomes an identity function, we obtain the Pareto optimality $f(A^*, x^*) = \min\{(f_2(A), f_3(x)), A \in \mathbf{A}, x \in X\}$. Using Pareto optimality is simpler, however, we lose an explicit dependence between several objectives (we keep a vector of objectives ignoring any priorities, on the other hand, we do not have problems combining objectives if they are measured in different "units", for example, an energy used and satisfaction of users).

For fixed $f_2$, we consider an optimization problem - looking for minimum of $f_3$, and for fixed $f_3$ we look for minimum of search costs - search optimum of $f_2$.

Objective functions allow capturing convergence and the convergence rate of construction of solutions much better than symbolic goals. Obviously every symbolic goal/termination condition can be expressed as an objective function. For example, a very simple objective function can be the following: if the goal is satisfied the objective is set to 1, and if not to 0. To reach such a goal is a maximization problem. Typically, much more complex objective functions are used to better express evolutions of solutions.

Let $(\mathbf{A}^*, X^*)$ denote the set of totally optimal solutions. In particular $X^*$ denotes the set of optimal solutions, and $\mathbf{A}^*$ the optimal search algorithms.

Let $Y$ be a metric space, where for every pair of its elements $x$ and $y$, there is assigned the real number $D(x, y) \geq 0$, called *distance*, satisfying three conditions (Kuratowski, 1977):

- $D(x, x) = 0$
- $D(x, y) = D(y, x)$
- $D(x, y) + D(y, z) \geq D(x, z)$

The distance function can be defined in different ways, e.g., as the Hamming distance, Euclidean distance, $D(x, y) = 0$ if x satisfies termination condition and $D(x, y) = 1$ otherwise.

To keep it independent from representation, and to allow to compare different solving algorithms, we will fix the distance function to the absolute value of difference of the objective functions $D(x, y) = |f(x) - f(y)|$. We extend the definition of the distance from the pairs of points to the distance between a point and the set of points $D(x, Y) = \inf\{|f(x) - f(y)|, y \in Y\}$.

In problem solving, we will be interested in the distance to the set of optimal solutions $Y^*$, i.e., in the distance $D((A, x), (\mathbf{A}^*, X^*))$, and in particular, in the distances $D(x, X^*)$ and $D(A, \mathbf{A}^*)$.

**Definition 2.6.** (*Solution convergence*) If there is a moment of time t (bounded or unbounded, i.e., $t = 0, 1, 2, 3, ...$) the solution will be said to be

- *convergent* to the total optimum iff there exists such $\tau$ that for every $t > \tau$, we have $D((A[t], X[t]), (\mathbf{A}^*, X^*)) = 0$,

- *divergent*, otherwise.

- *asymptotically convergent* to the total optimum iff for every $\varepsilon$, $\infty > \varepsilon > 0$, there exists such $\tau$ that for every $t > \tau$, we have $D((A[t], X[t]), (\mathbf{A}^*, X^*)) < \varepsilon$,

- *asymptotically divergent*, otherwise.

- *convergent with an error r* to the total optimum, where $\infty > r > 0$ iff there exists such such $\tau$ that for every $t > \tau$, we have $D((A[t], X[t]), (\mathbf{A}^*, X^*)) \leq r$,

- *fuzzy divergent*, otherwise.

Solution convergence corresponds to topological convergence in discrete spaces. Asymptotical solution convergence corresponds to topological convergence in metric spaces (Kelly, 1957). Solution convergence with an error corresponds to fuzzy convergence in metric spaces (Burgin, 2000).

If *t* is fixed, the convergence is *recursive*, otherwise it is *superrecursive*. Asymptotic convergence is superrecursive.

**Definition 2.7.** (*Solution convergence rate*) The *convergence rate* to the total optimum is defined as $D((A[t], X[t]), (\mathbf{A}^*, X^*)) - D((A[t+1], X[t+1]), (\mathbf{A}^*, X^*))$.

As in the case of differentiable functions, the solution convergence rate is the derivative of the distance function $D((A[t], X[t]), (A^*, X^*))$. Only it is not the conventional derivative but a fuzzy derivative in the sense of (Burgin, 2001). In contrast to the conventional differentiation, fuzzy differentiation developed in neoclassical analysis allows one to differentiate discontinuous and even discrete functions.

The convergence rate describes the one-step performance of the algorithm, where the positive convergence rate means that the algorithm drifts towards the optimum and the negative rate signifies a drift away from the optimum. With positive convergence rate, the search algorithm will typically converge or asymptotically converge to the optimum. The best search algorithms will have typically a high convergence rate and a small number of steps to reach the optimum. In the similar way, optimal and search optimal convergence and convergence rate can be defined. If the search algorithm is probabilistic, we use an expected value of the distance function.

Search can involve single or multiple agents:

- single agent: and algorithms like depth-first, breadth-first, uniform cost, iterative deepening, A*, IDA*, SMA*, hill climbing, simulated annealing (Russell, 1995, Michalewicz & Fogel, 2000),
- two agents: using algorithms like minimax, alpha-beta, expectiminimax (von Neumann & Morgenstern, 1944, Russell, 1995),
- multiple agents: and algorithms like $k\Omega$-optimization, *n*-player games, co-evolutionary algorithms, COllective INtelligence (Eberbach, 2005b; von Neumann & Morgenstern, 1944; Michalewicz & Fogel, 2000; Wolpert, 2000).

For multiple agents search can be *cooperative*, *competitive*, or *independent*. In *cooperative search*, agents use results of others to find an optimum; in *competitive search*, agents compete for resources for the search optimum, and in *independent search* agents do not interact.

In a case when the optimization space *X* consists of algorithms, there are two classes of search (selection) algorithms: online and offline. In *online* algorithms, action execution and computation are interleaved, while in *offline* algorithms, the complete solution for an optimization problem is computed first and executed after without any perception. More interesting are online algorithms, although the majority of developed so far search algorithms are offline.

## 3. Inductive and Limit Turing Machines

Here we give a very short description of inductive Turing machines, while a more detailed exposition is given in (Burgin, 2005). The structure of an inductive Turing machine, as an abstract automaton, consists of three components called *hardware*, *software*, and *infware*. Infware is a description and specification of information that is processed by an inductive Turing machine. Computer infware consists of data processed by the computer. Inductive Turing machines are abstract automata working with the same symbolic information in the form of words as conventional Turing machines. Consequently, formal languages with which inductive Turing machines works constitute their infware.

Computer hardware consists of all devices (the processor, system of memory, display, keyboard, etc.) that constitute the computer. In a similar way, an inductive Turing machine $M$ has three abstract devices: a *control device A*, which is a finite automaton and controls performance of $M$; a *processor* or *operating device H*, which corresponds to one or several *heads* of a conventional Turing machine; and the *memory E*, which corresponds to the *tape* or tapes of a conventional Turing machine. The memory $E$ of the simplest inductive Turing machine consists of three linear tapes, and the operating device consists of three heads, each of which is the same as the head of a Turing machine and works with the corresponding tapes.

The *control device A* is a finite automaton that regulates: the state of the whole machine $M$, the processing of information by $H$, and the storage of information in the memory $E$.

The *memory E* is divided into different but, as a rule, uniform cells. It is structured by a system of relations that organize memory as well-structured system and provide connections or ties between cells. In particular, *input* registers, the *working* memory, and *output* registers of $M$ are separated. Connections between cells form an additional structure *K* of *E*. Each cell can contain a symbol from an alphabet of the languages of the machine *M* or it can be empty.

In a general case, cells may be of different types. Different types of cells may be used for storing different kinds of data. For example, binary cells, which have type B, store bits of information represented by symbols 1 and 0. Byte cells (type BT) store information represented by strings of eight binary digits. Symbol cells (type SB) store symbols of the alphabet(s) of the machine *M*. Cells in conventional Turing machines have SB type. Natural number cells, which have type NN, are used in random access machines (Aho, et al, 1976). Cells in the memory of quantum computers (type QB) store q-bits or quantum bits (Deutsch, 1985). Cells of the tape(s) of real-number Turing machines (Burgin, 2005). have type RN and store real numbers. When different kinds of devices are combined into one, this new device has several types of memory cells. In addition, different types of cells facilitate modeling the brain neuron structure by inductive Turing machines.

It is possible to realize an arbitrary structured memory of an inductive Turing machine *M*, using only one linear one-sided tape *L*. To do this, the cells of *L* are enumerated in the natural order from the first one to infinity. Then *L* is decomposed into three parts according to the input and output registers and the working memory of *M*. After this, nonlinear connections between cells are installed. When an inductive Turing machine with this memory works, the head/processor is not moving only to the right or to the left cell from a given cell, but uses the installed nonlinear connections.

Such realization of the structured memory allows us to consider an inductive Turing machine with a structured memory as an inductive Turing machine with conventional tapes in which additional connections are established. This approach has many advantages. One of them is that inductive Turing machines with a structured memory can be treated as multitape automata that have additional structure on their tapes. Then it is conceivable to study different ways to construct this structure. In addition, this representation of memory allows us to consider any configuration in the structured memory *E* as a word written on this unstructured tape.

If we look at other devices of the inductive Turing machine *M*, we can see that the processor *H* performs information processing in *M*. However, in comparison to computers, this operational device performs very simple operations. When *H* consists of one unit, it can change a symbol in the cell that is observed by *H*, and go from this cell to another using a connection from **K**. This is exactly what the head of a Turing machine does.

It is possible that the processor *H* consists of several processing units similar to heads of a multihead Turing machine. This allows one to model in a natural way various real and abstract computing systems by inductive Turing machines. Examples of such systems are: multiprocessor computers; Turing machines with several tapes; networks, grids and clusters of computers; cellular automata; neural networks; and systolic arrays.

We know that programs constitute computer software and tell the system what to do (and what not to do). The *software* **R** of the inductive Turing machine *M* is also a program in the form of simple rules:

$q_h a_i \rightarrow a_j q_k$ 	(1)

$q_h a_i \rightarrow c q_k$ 	(2)

$q_h a_i \rightarrow a_j q_k c$ 	(3)

Here $q_h$ and $q_k$ are states of *A*, $a_i$ and $a_j$ are symbols of the alphabet of *M*, and *c* is a type of connection in the memory *E*.

Each rule directs one step of computation of the inductive Turing machine *M*. The rule (1) means that if the state of the control device *A* of *M* is $q_h$ and the processor *H* observes in the cell the symbol $a_i$, then the state of *A* becomes $q_k$ and the processor *H* writes the symbol $a_j$ in the cell where it is situated. The rule (2) means that the processor *H* then moves to the next cell by a connection of the type *c*. The rule (3) is a combination of rules (1) and (2).

Like Turing machines, inductive Turing machines can be deterministic and nondeterministic. For a *deterministic* inductive Turing machine, there is at most one connection of any type from any cell. In a *nondeterministic* inductive Turing machine, several connections of the same type may go from some cells, connecting it with (different) other cells. If there is no connection of the prescribed by an instruction type that goes from the cell that is observed by *H*, then *H* stays in the same cell. There may be connections of a cell with itself. Then *H* also stays in the same cell. It is possible that *H* observes an empty cell. To represent this situation, we use the symbol ε. Thus, it is possible that some elements $a_i$ and/or $a_j$ in the rules from **R** are equal to ε in the rules of all types. Such rules describe situations when $H$ observes an empty cell and/or when $H$ simply erases the symbol from some cell, writing nothing in it.

The rules of the type (3) allow an inductive Turing machine to rewrite a symbol in a cell and to make a move in one step. Other rules (1) and (2) separate these operations. Rules of the inductive Turing machine $M$ define the transition function of $M$ and describe changes of $A$, $H$, and $E$. Consequently, they also determine the transition functions of $A$, $H$, and $E$.

A general step of the machine $M$ has the following form. At the beginning of any step, the processor $H$ observes some cell with a symbol $a_i$ (for an empty cell the symbol is Λ) and the control device $A$ is in some state $q_h$.

Then the control device $A$ (and/or the processor $H$) chooses from the system **R** of rules a rule $r$ with the left part equal to $q_h a_i$ and performs the operation prescribed by this rule. If there is no rule in **R** with such a left part, the machine $M$ stops functioning. If there are several rules with the same left part, $M$ works as a nondeterministic Turing machine, performing all possible operations. When $A$ comes to one of the final states from $F$, the machine $M$ also stops functioning. In all other cases, it continues operation without stopping.

For an abstract automaton, as well as for a computer, three things are important: how it receives data, process data and obtains its results. In contrast to Turing machines, inductive Turing machines obtain results even in the case when their operation is not terminated. This results in essential increase of performance abilities of systems of algorithms.

The computational result of the inductive Turing machine $M$ is the word that is written in the output register of $M$: when $M$ halts while its control device $A$ is in some final state from $F$, or when $M$ never stops but at some step of computation the content of the output register becomes fixed and does not change although the machine $M$ continues to function. In all other cases, $M$ gives no result.

**Definition 3.1.** The memory $E$ is called *recursive* if all relations that define its structure are recursive.

Here recursive means that there are some Turing machines that decide/build all naming mappings and relations in the structured memory.

**Definition 3.2.** Inductive Turing machines with recursive memory are called *inductive Turing machines of the first order*.

**Definition 3.3.** The memory $E$ is called *n-inductive* if all relations that define its structure are constructed by an inductive Turing machine of order $n$.

**Definition 3.4.** Inductive Turing machines with $n$-inductive memory are called *inductive Turing machines of the order $n + 1$*.

Limit Turing machines have the same structure (hardware) as inductive Turing machines. The difference is in a more general way in obtaining the result of computation. To obtain their result, limit Turing machines need some topology in the set of all words that are processed by these machines.

Let a limit Turing machine $L$ works with words in an alphabet $A$ and in the set $A^*$ of all such words, a topology $T$ is defined. While the machine $L$ works, it produces in the output tape (memory) words $w_1, w_2, \ldots, w_n, \ldots$ . Then the result of computation of the limit Turing machine $L$ is the limit of this sequence of words in the topology $T$.

When the set $A^*$ has the discrete topology, limit Turing machines coincide with Turing machines.

## 4. Evolutionary Algorithms and Evolutionary Turing Machines

An evolutionary algorithm is a probabilistic beam hill climbing search algorithm directed by the fitness objective function. The beam (population size) maintains multiple search points, hill climbing means that only a current search point from the search tree is remembered, and a termination condition very often is set to the optimum of the fitness function.

**Definition 4.1.** A generic *evolutionary algorithm (EA)* can be described in the form of the functional equation (recurrence relation) working in a simple iterative loop in discrete time t, called generations, t = 0, 1, 2,... (Fogel, 1995, Michalewicz & Fogel, 2004, Fogel, 2001):

$X[t+1] = s (v (X[t]))$, where

- $X[t] \subseteq X$ is a *population* under a representation consisting of one or more individuals from the set $X$ (e.g., fixed binary strings for genetic algorithms (GAs), Finite State Machines for evolutionary programming (EP), parse trees for genetic programming (GP), vector of reals for evolution strategies (ES)),

- $s$ is a *selection* operator (e.g., *truncation, proportional, tournament*),

- $v$ is a *variation* operator (e.g., variants of *mutation* and *crossover*),

- $X[0]$ is an *initial* population,

- $F \subseteq X$ is the set of *final* populations satisfying the *termination condition* (goal of evolution). The desirable termination condition is the optimum in $X$ of the *fitness* function $f: X \rightarrow \mathbf{R}$, which is extended to the *fitness* function $f(X[t])$ of the best individual in the population $X[t] \in F$, where $f$ is defined typically in the domain of nonnegative real numbers. In many cases, it is impossible to achieve or verify this optimum. Thus, another stopping criterion is used (e.g., the maximum number of generations, the lack of progress through several generations.).

Definition 4.1 is applicable to all typical EAs, including GA, EP, ES, GP. It is possible to use it to describe other emerging subareas like ant colony optimization, or particle swarm optimization. Co-evolutionary algorithms use typically multiple populations, e.g., vectors of representation vectors are evolved. In fact, there is no restriction on the type of representation used. Sometimes only the order of variation and selection is reversed, i.e., selection is applied first, and variation second. Variation and selection depend on the fitness function. Of course, it is possible to think and implement more complex variants of evolutionary algorithms.

Evolutionary algorithms evolve population of solutions x, but they may be the subject of self-adaptation (like in ES) as well. This extension has been used in Evolution Strategies (although typically limited only to ES parameter optimization, e.g., evolution of standard deviation in Gaussian mutation). Technically, it means that the domain of the variation operator $v$, selection operator $s$, and the fitness function $f$ are extended to operate both on the population under representation $x$ as well as on the encoding of the evolutionary algorithm. In the next part of this paper, we discuss this more general EC evolving in parallel its population $x$, as well as an evolutionary algorithm (perhaps, both evolved using different time scales). For sure, evolution in nature is not static, the rate of evolution fluctuates, their variation operators are subject to slow or fast changes, its goal (if it exists at all) can be a subject of modifications as well.

For the readers who would argue that most EC applications use currently static evolutionary algorithms, our approach will still be valid by assuming that the utilized evolutionary algorithm is fixed. The advantage of non-static evolutionary algorithms is that they allow capturing the complexity and adaptation of the search process.

Formally, an evolutionary algorithm looking for the optimum of the fitness function violates some classical requirements of recursive algorithms. If its termination condition is set to the optimum of the fitness function, it may not terminate after a finite number of steps. To fit it to the old "algorithmic" approach, an artificial (or somebody can call it pragmatic) stop criterion has had to be added (see e.g., (Michalewicz, 1996; Koza, 1992)). The evolutionary algorithm, to remain recursive, has to be stopped after a finite number of generations or when no visible progress is observable. Naturally, in a general case, Evolutionary Algorithms are instances of super-recursive algorithms.

Now, we define a formal algorithmic model of Evolutionary Computation - an *Evolutionary Turing Machine* (Eberbach, 2005a).

**Definition 4.2.** An *evolutionary Turing machine* (*ETM*) $E = \{$ TM$[t]$; $t = 0, 1, 2, 3, ... \}$ is a series of (possibly infinite) Turing machines TM$[t]$ each working on population $X[t]$ in generations $t = 0, 1, 2, 3, ...$ where

- each $\delta[t]$ transition function (rules) of the Turing Machine TM$[t]$ represents (encodes) an evolutionary algorithm that works with the population $X[t]$, and evolved in generations $0, 1, 2, ... , t$,

- only generation 0 is given in advance, and any other generation depends on its predecessor only, i.e., the outcome of the generation $t = 0, 1, 2, 3, ...$ is the population $X[t + 1]$ obtained by applying the recursive variation $v$ and selection $s$ operators working on population $X[t]$,

- (TM$[0]$, $X[0]$) is the initial Turing Machine operating on its input - an initial population $X[0]$,

- the goal (or halting) state of ETM $E$ is represented by any population $X[t]$ satisfying the termination condition. The desirable termination condition is the optimum of the fitness performance measure $f(x[t])$ of the best individual from the population $X[t]$.

- When the termination condition is satisfied, then the ETM $E$ halts ($t$ stops to be incremented), otherwise a new input population $X[t + 1]$ is generated by TM[$t + 1$].

**Remark 4.1.** Turing machines TM[$t$] perform multiple computations in the sense of (Burgin, 1983).

**Remark 4.2.** Variation and selection operators are recursive to allow problem computation on Turing machines. Later we will release that restriction to allow nonrecursive solutions.

**Remark 4.3.** We do not consider here such ETM that change transition functions $\delta[t]$ and/or memory of the Turing machines TM[$t$] or/and fitness functions. We study these machines in another work.

**Remark 4.4.** In general, because the fitness function can be the subject of evolution as well, evolution is potentially an *infinite* process. Changing the transition function $\delta[t]$ of the TM can be thought as some kind of evolvable hardware, or assuming fixed hardware we can think about reprogrammable evolutionary algorithms.

In this model, both variation $v$ and selection $s$ operators are realized by Turing machines. So, it is natural that the same Turing machine computes values of the fitness function $f$. This brings us to the concept of a weighted Turing machine.

**Definition 4.3.** An *weighted Turing machine* $(T, f)$ computes a pair $(x, f(x))$ where $x$ is a word in the alphabet of $T$ and $f(x)$ is the value of the evaluation function $f$ of the machine $(T, f)$.

Examples of weighted Turing machines are fuzzy Turing machines (Wiedermann, 2004), which are theoretical model for fuzzy algorithms (Zadeh, 1968; Zheru Chi, et al, 1996).

Another example of weighted Turing machines in particular and weighted algorithms in general are Turing machines that compute recursive real numbers and recursive real-valued functions (Rice, 1951; Freund, 1983).

Weighted algorithms find applications in many areas (cf., for example, (JiJi, et al, 2000) for chemistry or (Arya, et al, 2001) for planar point location).

It is necessary to remark that only in some cases it is easy to compute values of the fitness function *f*. Examples of such situations are such fitness functions as the length of a program or the number of parts in some simple system. However, in many other cases, computation of the values of the fitness function *f* can be based on a complex algorithm and demand many operations. For instance, when the optimized species are programs and the fitness function *f* is time necessary to achieve the program goal, then computation of the values of the fitness function *f* can demand functioning or simulation of programs generated in the evolutionary process. We encounter similar situations when optimized species are computer chips or parts of plane or cars. In this case, computation of the values of the fitness function *f* involves simulation.

Weighted computation realized by weighted Turing machines allows us to extend the formal algorithmic model of Evolutionary Computation defining a *Weighted Evolutionary Turing Machine*.

**Definition 4.4.** A *weighted evolutionary Turing machine* (*WETM*) $E = \{ TM[t]; t = 0, 1, 2, 3, ... \}$ is a series of (possibly infinite) weighted Turing machines TM[$t$] each working on population $X[t]$ in generations $t = 0, 1, 2, 3, ...$ where

- each $\delta[t]$ transition function (rules) of the weighted Turing machine TM[$t$] represents (encodes) an evolutionary algorithm that works with the population $X[t]$, and evolved in generations $0, 1, 2, ... , t$,

- only generation 0 is given in advance, and any other generation depends on its predecessor only, i.e., the outcome of the generation $t = 0, 1, 2, 3, ...$ is the population $X[t + 1]$ obtained by applying the recursive variation $v$ and selection $s$ operators working on population $X[t]$ and computing the fitness function *f*,

- (TM[0], $X[0]$) is the initial weighted Turing Machine operating on its input - an initial population $X[0]$,

- the goal (or halting) state of WETM *E* is represented by any population $X[t]$) satisfying the termination condition. The desirable termination condition is the optimum of the fitness performance measure $f(x[t])$ of the best individual from the population $X[t]$.

- When the termination condition is satisfied, then the WETM *E* halts (*t* stops to be incremented), otherwise a new input population $X[t + 1]$ is generated by TM$[t + 1]$.

The concept of a universal automaton/algorithm plays an important role in computing and is useful for different purposes. In the most general form this concept is developed in (Burgin, 2005a).

The construction of universal automata and algorithms is usually based on some codification (symbolic description) **c**: **K** $\to$ *X* of all automata/algorithms in **K**.

**Definition 4.5.** An automaton/algorithm *U* is *universal* for the class **K** if given a description **c**(*A*) of an automaton/algorithm *A* from **K** and some input data *x* for it, *U* gives the same result as *A* for the input *x* or gives no result when *A* gives no result for the input *x*.

This leads us immediately, following Turing's ideas, to the concept of the universal Turing machine and its extensions - a *Universal Evolutionary Turing Machine* and *Weighted Evolutionary Turing Machine*. We can define a Universal Evolutionary Turing Machine as an abstraction of all possible ETMs, in the similar way, as a universal Turing machine has been defined, as an abstraction of all possible Turing machines.

Let **A** be an optimizing (algorithm) space with the optimization space *X* and **c**: **A** $\to$ *I* be a codification (symbolic description) of all automata/algorithms in **A**. Evolutionary algorithms are series of algorithms from **A**. For instance, an evolutionary Turing machine is a series of Turing machines.

**Definition 4.6.** A *universal evolutionary Turing machine* (*UETM*) is an ETM *EU* with the optimization space $Z = X \times I$. Given a pair ( **c**(*E*), *X*[0]) where *E* = { TM[*t*]; *t* = 0, 1, 2, 3, ... } is an ETM and *X*[0] is the start population, the machine *EU* takes this pair as its input and produces the same population *X*[1] as the Turing machine TM[0] working with the same population *X*[0]. Then *EU* takes the pair ( **c**(*E*), *X*[1]) as its input and produces the same population *X*[2] as the Turing machine TM[1] working with the population *X*[1]. In general, *EU* takes the pair ( **c**(*E*), *X*[*t*]) as its input and produces the same population $X[t + 1]$ as the Turing machine TM[*t*] working with the population *X*[*t*] where *t* = 0, 1, 2, 3, ... .

In other words, by a *Universal Evolutionary Turing Machine* (*UETM*) we mean such ETM $U$ that on each step takes as the input a pair ( **c**(TM[$t$]), $X[t]$) and behaves like ETM $E$ with input $X[t]$ for $t = 0, 1, 2, ....$ UETM $U$ stops when ETM $E$ stops.

Definition 4.6 gives properties of but does not imply its existence. However, as in the case of Turing machines, we have the following result.

**Theorem 4.1** (Eberbach, 2005). In the class of all evolutionary Turing machines, there is a universal universal evolutionary Turing machine.

Using the structure of the universal Turing machine, we can get an explicit construction of a universal evolutionary Turing machine.

It is possible to build a *universal evolutionary Turing machine* (*UETM*) as a series $EU = \{ UT[t]; t = 0, 1, 2, 3, ... \}$ of (possibly infinite) instances of Universal Turing machines UT[$t$] working on pairs ( **c**(TM[$t$]), $X[t]$) in generations $t = 0, 1, 2, 3, ... $, where

- each TM[t] represents (encodes) the component $t$ of the evolutionary algorithm $E$ with population $X[t]$, and evolved in generations 0,1,2,...,t,

- only generation 0 is given in advance, and any other generation depends on its predecessor only, i.e., the outcome of generation t = 0, 1, 2, ... is the pair (TM[$t$+1], $X[t$ +1]) by applying the recursive variation v and selection s operators operating on population x and (possibly) on evolutionary algorithm M as well,

- UT[0] is the initial evolutionary algorithm operating on its input - an initial population $X[0]$,

- the goal (or halting) state of UETM is represented by any pair (TM[t], $X[t]$) satisfying the termination condition. The desirable termination condition is the optimum of the fitness performance measure $f(M[t], X[t]) = f_1(f_2(M[t]), f_3(X[t]))$ of the best individual from the population of solutions and evolutionary algorithms, where $f_1$ is an aggregating function, $f_2$ is an evolutionary algorithm fitness function, and $f_3$ is a problem-specific fitness function. If the termination condition is satisfied, then the UETM halts (*t* stops to be incremented), otherwise a new pair TM[$t$ + 1] and its input/population $X[t$ +1] is generated.

Note that an infinite sequence of Turing machines in ETM (UETM) generally may work like the limit Turing machine. The limit Turing machine is more expressive than Turing machine, thus evolutionary Turing machines is more expressive than Turing machines, i.e., it belongs to superTuring models of computation (Eberbach, Goldin & Wegner, 2004).

**Definition 4.7.** A *universal weighted evolutionary Turing machine* (*UWETM*) is an WETM *EU* with the optimization space $Z = X \times I$. Given a pair ( **c**(*E*), *X*[0]) where *E* = { TM[*t*]; *t* = 0, 1, 2, 3, ... } is an WETM and *X*[0] is the start population, the machine *EU* takes this pair as its input and produces the same population *X*[1] as the weighted Turing machine TM[0] working with the same population *X*[0]. Then *EU* takes the pair ( **c**(*E*), *X*[1]) as its input and produces the same population *X*[2] as the weighted Turing machine TM[1] working with the population *X*[1]. In general, *EU* takes the pair ( **c**(*E*), *X*[*t*]) as its input and produces the same population *X*[*t* + 1] as the weighted Turing machine TM[*t*] working with the population *X*[*t*] where *t* = 0, 1, 2, 3, ... .

This definition gives properties of but does not imply its existence.

**Theorem 4.2.** In the class of all weighted evolutionary Turing machines with a given recursively computable weight (fitness) function *f*, there is a universal weighted evolutionary Turing machine.

**Theorem 4.3.** In the class of all weighted evolutionary Turing machines with recursively computable weight (fitness) functions, there is a universal weighted evolutionary Turing machine.

## 5. Types of Evolutionary Computations

We know that the same hardware allows the computer to realize/use different modes of computation. In a similar way, the same algorithmic structure provides for different types of evolutionary computations.

Three finite modes (types) of evolutionary computations:

1. *Bounded finite evolutionary computations* when there are only finite numbers of *TM*[*t*] and *t* < *C*.
2. *Unbounded finite* (*potentially infinite*) *evolutionary computations* when at each moment of time there are only finite numbers of *TM*[*t*].

3. *Infinite evolutionary computations* when it is possible that at some moments of time there is an infinite number of *TM*[*t*].

These types have specific computational subtypes.

Three bounded finite modes (types) of evolutionary computations:

1a. *Recursive*: each machine *TM*[*t*] gives the final result after a finite number of steps and after this stops the process of computation or, at least, the machine informs when the result is obtained.

1b. *Inductive*: each machine *TM*[*t*] gives the final result after a finite number of steps but it does not always after this stop the process of computation or informs when the result is obtained.

1c. *Limit*: each machine *TM*[*t*] gives the partial result after a finite number of steps and the final result is the limit of these partial results.

**Example 5.1.** Evolutionary algorithms with the termination condition set to the fixed maximum number of generations belong to the class 1a. In general, such evolutionary algorithms improve solutions, but do not guarantee to find a global optimum of fitness function.

**Example 5.2.** Class 1b can be represented by evolutionary algorithms with unknown or very complex fitness function, where global optimum can be hit, but we are unable to verify that and next generation is invoked. Thus in theory, the computation can last forever, however global optimum if found is received in a finite number of steps (we put limit on the number of generations).

**Example 5.3.** Class 1c can be represented by evolutionary algorithms with elitism, completeness, and looking for optimum of the fitness function. In many cases (small search spaces or we are lucky), the last generation will contain the optimal solution.

Three unbounded finite modes (types) of evolutionary computations:

2a. *Recursive*: the sequence of machines *TM*[*t*] gives the final result after a finite number of steps and after this stops the process of computation or, at least, one of the machines *TM*[*t*] informs when the result is obtained.

2b. *Inductive*: the sequence of machines *TM*[*t*] gives the final result after a finite number of steps but it does not always after this stop the process of computation or informs when the result is obtained.

2c. *Limit*: results of machines *TM*[*t*] are only partial and the final result is the limit of these partial results.

**Example 5.4.** Class 2a can be represented by evolutionary algorithms with termination condition set to fixed but unbounded number of generations. Another system that belongs to the class 2a are evolutionary algorithms with termination condition set to the lack of improvement of the fitness function – such process is potentially unbounded.

**Example 5.5.** Class 2b can be represented by evolutionary algorithms with very complex or unknown fitness function where the number of generations is fixed but unbounded. Once again, the optimum can be reached, but there is no guarantee that it will be maintained (can be not recognized and lost). However, the process of search is fixed, but potentially infinite.

**Example 5.6.** Class 2c can be represented by evolutionary algorithms with elitism, completeness and looking for optimum of the fitness function in unbounded but fixed number of generations. The limit TM will guarantee to contain optimal solution (potentially in infinity).

**Example 5.7.** Evolutionary algorithms with elitism, completeness and looking for the best evolutionary algorithm belong to the class 3. As we know, the best evolutionary algorithm does not exist, i.e., the limit solution is outside of domain of evolutionary algorithms. However, this outside of domain limit (asymptote), can be approximated by an infinite sequence of better evolutionary algorithms, but the process will never stops and will require an infinite number of Turing machines.

# 6. Completeness, Optimality, Search Optimality, Total Optimality and Decidability Results

In this section, we present results on completeness, optimality and decidability of classes of evolutionary algorithms introduced in Section 5.

**Definition 6.1.** (*Completeness of evolutionary algorithms search*) Evolutionary computation search is *complete* if UETM starting from its initial state (*TM*[0], *X*[0]) using its variation and selection operators guarantees to reach state $(TM[t], X[t])$ satisfying the termination condition, pending there is one.

**Definition 6.2.** (*Optimality of evolutionary search*) Evolutionary computation search is *optimal* if a UETM has its termination condition set to the optimum of the problem-specific fitness function and $f_3(X[t])$ is convergent (or asymptotically convergent) to the set of optimal solutions $X^*$.

**Definition 6.3.** (*Search optimality of evolutionary algorithms*) Evolutionary computation is *search optimal* if a UETM has its termination condition set to the optimum of the search fitness function together with problem-specific termination condition allowing to find any solution $X[t]$, and $f_2(TM[t])$ is convergent (or asymptotically convergent) to the set of optimal solutions $\mathbf{A}^*$.

**Definition 6.4.** (*Total optimality of evolutionary search*) Evolutionary computation search is *totally optimal* if UETM has its termination condition set to the optimum of the fitness function $f(TM[t], X[t])$ and $f$ is convergent (or asymptotically convergent) to the set of optimal solutions $(\mathbf{A}^*, X^*)$.

**Definition 6.5.** (*Decidability of evolutionary search*) Evolutionary computation search is:

1) *recursively decidable* if problems of completeness, optimality, search optimality, total optimality can be solved in a finite number of steps with process termination;

2) *inductively decidable* if problems of completeness, optimality, search optimality, total optimality can be solved in a finite number of steps;

3) *asymptotically decidable* if problems of completeness, optimality, search optimality, total optimality can be solved in the limit.

**Theorem 6.1.** Bounded finite evolutionary computation

- is complete if the termination condition is set to the fixed number of generations and may be incomplete if it is set to the optimum of fitness function;

- is not guaranteed to be optimal (search optimal, or totally optimal) for the termination condition set to the optimum of the fitness function $f_3$, ($f_2$, or $f$, respectively);

- is recursively undecidable.

**Theorem 6.2.** Unbounded finite evolutionary computation

- is complete if the termination condition is set to the fixed number of generations and may be incomplete if it is set to the optimum of fitness function;
- is optimal (search optimal, or totally optimal) for the termination condition set to the optimum of the fitness function $f_3$, ($f_2$, or $f$, respectively) if it is complete and elitism selection is used;
- is recursively undecidable for the search for the best evolutionary algorithm;
- is inductively decidable for the search for the best evolutionary algorithm if each generation is generated by a recursive algorithm, e.g., Turing machine.

**Theorem 6.3.** Infinite evolutionary computation

- is complete for the termination condition set to the optimum of the fitness function if the transition probability for any pair of states (M, $x$), (N, $y$) and time $\tau > t \geq 0$ satisfies $P((M[t],x[t]),(N[\tau],y[\tau])) > 0$;
- is optimal (search optimal, or totally optimal) for the termination condition set to the optimum of the fitness function $f_3$, ($f_2$, or $f$, respectively) if it is complete and elitism selection is used;
- is decidable in the limit if complete and elitism selection is used (including search for the best evolutionary algorithm).

## 7. Cooperation and Competition in Evolutionary Computation

Popular models of distributed intelligent performance (e.g., optimization) are coevolutionary systems, Particle Swarm Optimization (PSO, also called Swarm Intelligence), and Ant Colony Optimization (ACO also known as Ant Colony System (ACS)).

In coevolutionary systems, see e.g., (Michalewicz and Fogel, 2004) more than one evolution process takes place: usually there are different populations, which *interact* with each other. In coevolutionary systems, being a special case of concurrent systems, the fitness function for one population may depend on the state of the evolution processes in the other population(s). This is important topic for modeling artificial life, some business applications, intelligent agents, games, etc.

Particle Swarm Optimization (PSO, also called Swarm Intelligence) is a multiagent/coevolutionary technique, developed by Jim Kennedy and Russell Eberhart in 1995 (Kennedy and Eberhart, 1995; Kennedy, et al., 2001). A population of particles "flies" through the problem space. PSO has been inspired by bird flocking, fish schooling, buffalo herds, and swarming theory of social behaviors. A bird flock becomes swarm steering toward the center, matching neighbors' velocity, and avoiding collisions. PSO operates on a population of individuals where variation is applied, but without selection. Each individual has a position and velocity that is updated according to the relationship between the individual's parameters and the best location of the individual in the population found so far. The search is biased toward better regions of space, with the result being a sort of "flocking" toward the best solutions. In Particle Swarm Optimization, the population always (on each step) consists of the same species (individual systems). What is changing are characteristics of these species, for example, their positions in the search space. In our model, these changes are computed by an ETM. This allows optimizing system to use not only linear shifts of particles in their search for the best position, but also more efficient recursive transformations of particles movements.

Ant Colony Optimization (ACO also known as Ant Colony System (ACS)) is another multiagent technique, developed by Marco Dorigo and his coauthors in 1997 (Bonabeau et al, 1999), where low-level interactions between artificial (i.e., simulated) ants result in a complex behavior of the larger ant colony. Social insects - ants, bees, termites, and wasps - can be viewed as powerful problem-solving systems with sophisticated collective intelligence. Composed of simple interacting agents, this intelligence lies in the networks of interactions among individuals and between individuals and the environment. Social insects find food, divide labor among nestmates, build nests, and respond to external challenges. ACO algorithms were inspired by colonies of real ants that deposit a chemical substance (pheromone) on the ground. This substance influences the behavior of individual ants. The greater the amount of pheromone is on a particular path, the larger the probability that an ant will select that path. Artificial ants in ACO algorithms behave similarly. An abstract pheromone parameter fulfills a similar function like PSO position/velocity, or simulated annealing temperature. It forms a communication/ interaction channel between ants.

Coevolution, ant colony optimization and particle swarm optimization seem be potentially the most useful subareas of evolutionary computation for expressing interaction of multiple agents (in particular, to express their cooperation and competition). However, paradoxically in most current applications, these techniques are used to obtain optimal solutions for optimization of single agent behavior (in presence of other agents – members of population), and not for the optimization of group of agents trying to achieve a common goal (represented by joint fitness function). This is primarily because fitness function is optimized for a single individual from the population, and not for the population as a whole.

Now, we define a formal algorithmic model of Evolutionary Computation with cooperation and competition – a *Parallel Evolutionary Turing Machine*.

**Definition 7.1.** An *parallel evolutionary Turing machine* (*PETM*) $E = \{ TM_i[t]; t = 0, 1, 2, 3, ...; i \in I \}$ consists of a collection of series of (possibly infinite) Turing machines $TM_i[t]$ each working on population $X[t]$ in generations $t = 0, 1, 2, 3, ...$ where

- each $\delta_i[t]$ transition function (rules) of the Turing machine $TM_i[t]$ represents (encodes) an evolutionary algorithm that works with the whole generation $X[t]$ based on its own fitness performance measure $f_i(x[t])$, and evolved in generations $0, 1, 2, ... , t$,

- the whole generation $X[t]$ is the union of all subgenerations $X_i[t]$ obtained by all Turing machines $TM_i[t - 1]$ that collaborate in generating $X[t]$,

- only the zero generation $X[0]$ is given in advance, and any other generation depends on its predecessor only, i.e., the outcome of the generation $t = 0, 1, 2, 3, ...$ is the subgeneration $X_i[t + 1]$ obtained by applying the recursive variation $v$ and selection $s$ operators working on the whole generation $X[t]$ and realized by the Turing machine $TM_i[t]$,

- $TM_i[0]$ are the initial Turing machines operating on its input - an initial population $X[0]$,

- the goal (or halting) state of PETM *E* is represented by any population *X*[*t*]) satisfying the termination condition. The desirable termination condition is the optimum of the unified fitness performance measure *f*(*X*[*t*]) of the whole population *X*[*t*].

- when the termination condition is satisfied, then the PETM *E* halts (*t* stops to be incremented), otherwise a new input population *X*[*t* + 1] is generated by machines TM$_i$[*t* + 1].

In a similar way, we define a Parallel Weighted Evolutionary Turing Machine.

**Definition 7.2.** A *parallel weighted evolutionary Turing machine* (*PWETM*) *E* = *E* = { TM$_i$[*t*]; *t* = 0, 1, 2, 3, ...; *i* ∈ *I* } consists of a collection of series of (possibly infinite) Turing machines TM$_i$[*t*] each working on population *X*[*t*] in generations *t* = 0, 1, 2, 3, ... where

- each δ$_i$[*t*] transition function (rules) of the Turing machine TM$_i$[*t*] represents (encodes) an evolutionary algorithm that works with the whole generation *X*[*t*] based on its own fitness performance measure *f$_i$*(*x*[*t*]), and evolved in generations 0, 1, 2, ... , *t*,

- the whole generation *X*[*t*] is the union of all subgenerations *X$_i$*[*t*] obtained by all Turing machines TM$_i$[*t* - 1] that collaborate in generating *X*[*t*],

- only the zero generation *X*[0] is given in advance, and any other generation depends on its predecessor only, i.e., the outcome of the generation *t* = 0, 1, 2, 3, ... is the subgeneration *X$_i$*[*t* + 1] obtained by applying the recursive variation *v* and selection *s* operators working on the whole generation *X*[*t*] and computing the fitness function *f$_i$* , and realized by the Turing machine TM$_i$[*t*],

- TM$_i$[0] are the initial Turing machines operating on its input - an initial population *X*[0],

- the goal (or halting) state of PETM *E* is represented by any population *X*[*t*]) satisfying the termination condition. The desirable termination condition is the optimum of the unified fitness performance measure *f*(*X*[*t*]) of the whole population *X*[*t*].

when the termination condition is satisfied, then the PETM $E$ halts ($t$ stops to be incremented), otherwise a new input population $X[t + 1]$ is generated by machines $TM_i[t + 1]$.

Our models (of PETM and PWETM) are already prepared to handle such situation. It is enough to assume that fitness functions $f$, $f_1$, $f_2$, and $f_3$ are computed for the whole population (perhaps, consisting of subpopulations), and not for separate individuals from the population only. Let us assume that our population $|x| = p$, i.e., it consists of $p$ individuals or subpopulations. For simplicity, let us consider only individuals (by adding multiple indices, we can consider subpopulations without losing the generality of the approach).

Let $f(M[t],x[t]) = f_1(f_2(M[t]),f_3(x[t]))$, where $M[t] = \{M_1[t],…,M_p[t]\}$, $x[t]=\{x_1[t],…,x_p[t]\}$. We define a problem-specific fitness function $f_3$ for the whole population $f_3(x[t]) = f_{13}(f_{31}(x_1[t]),…,f_{3p}(x_p[t]))$, where $f_{13}$ is an aggregating function for $f_{31}, …, f_{3p}$, and $f_{3j}$ is a fitness function of the $j$-th individual $x_j$, $j = 1, …, p$, and an evolutionary algorithm fitness function $f_2$ for the whole population $f_2(M[t])=f_{12}(f_{21}(M_1[t]),…,f_{2p}(M_p[t]))$, where $f_{12}$ is an aggregating function for $f_{21},…,f_{2p}$, and $f_{2j}$ is a fitness function of the $j$-th evolutionary algorithm $M_j$, $j = 1, …, p$, and evolutionary algorithm $M_j$ is responsible for evolution of $x_j$.

We will present definition for cooperation and competition for problem-specific fitness function $f_3$. Similar definitions can be provided for fitness functions $f$ and $f_2$.

**Definition 7.3.** (*Cooperation of single individual with population*) We will say that $j$-th individual *cooperates* in time t with the whole population on problem specific fitness function $f_3$ iff $f_3[t] > f_3[t+1]$ and other's individuals fitness functions are fixed, i.e., $f_{3i}[t] = f_{3i}[t+1]$ for $i \neq j$.

**Definition 7.4.** (*Cooperation of the whole population*) We will say that all population *cooperates* as the whole in time t on problem specific fitness function $f_3$ iff $f_3[t] > f_3[t+1]$.

**Definition 7.5.** (*Competition of single individual with population*) We will say that $j$-th individual *competes* in time t with the whole population on problem specific fitness function $f_3$ iff $f_3[t] < f_3[t+1]$ and other's individuals fitness functions are fixed $f_{3i}[t] = f_{3i}[t+1]$ for $i \neq j$.

**Definition 7.6.** (*Competition of the whole population*) We will say that all population *competes* as the whole in time t on problem specific fitness function $f_3$ iff $f_3[t] < f_3[t+1]$.

In other words, if individual improves (makes worse) fitness function of the whole population then it cooperates (competes) with it. If fitness function of the whole population improves (deteriorates) then the population exhibits cooperation (competition) as the whole (independently what its individuals are doing). If individual (population) cooperates (competes) for all moments of time, then it is always cooperative (competitive). Otherwise, it may sometimes cooperate, sometimes compete (like in Iterated Prisoner Dilemma problem).

Let us consider some problems.

*Analysis problem for $f_3$* : Given $f_{31}[0],\ldots,f_{3p}[0]$ for individuals $x_1[0],\ldots,x_p[0]$ from the population $X[0]$ and $f_{13}[0]$ is given. What will be the behavior (emerging, limit behavior) of $f_3[t]$ for $X[t]$?

*Synthesis/design problem for $f_3$* : Given $f_3[0]$ for the population $X[0]$. Find corresponding individuals $x_1[0],\ldots,x_p[0]$ with $f_{31}[0],\ldots,f_{3p}[0]$ and $f_{13}[0]$ that $f_3[t]$ will converge to optimum.

**Theorem 7.1.** (*Optimality of evolutionary computation with cooperating population – sufficient conditions to solve the synthesis problem for $f_3$*) For a given evolutionary algorithm $UT[0]$ with population $X[0]$, if UETM $EU = \{ UT[t]; t = 0, 1, 2, 3, ... \}$ satisfies three conditions;

1. the termination condition is set to the optimum of the problem-specific fitness function $f_3(X[t])$ with the optimum $f_3^*$,

2. search is complete, and

3. population is cooperative all time $t = 0, 1, 2, \ldots$,

then UETM $EU$ is guaranteed to find the optimum $X^*$ of $f_3(X[t])$ in an unbounded number of generations $t = 0, 1, 2, \ldots$, and that optimum will be maintained thereafter.

Note that cooperation replaces elitism in sufficient condition for convergence of cooperating members of population looking for the optimum of fitness of the whole population and not of the separate individual. There is no surprise: if the whole population competes all the time, then the optimum will not be found despite completeness.

**Theorem 7.2.** (*Optimality of evolutionary computation with competing population – inability to solve the synthesis problem for $f_3$*) For a given evolutionary algorithm *UT*[0] with population *X*[0], if UETM *EU* = { UT[*t*]; *t* = 0, 1, 2, 3, ... } satisfies three conditions

1. the termination condition is set to the optimum of the problem-specific fitness function $f_3(X[t])$ with the optimum $f_3^*$,

2. search is complete, and

3. population is competing all time $t$ = 0, 1, 2, … ,

then UETM *EU* is not guaranteed to find the optimum $X^*$ of $f_3(X[t])$ even in an unbounded number of generations $t$ = 0, 1, 2, ....

If population is sometimes competing, sometimes cooperating, then the optimum sometimes will be found, sometimes not, but the convergence and its maintenance is not guaranteed.

Analogous results can be derived for search optimization and total optimization problems.

# 8. Conclusions

In this paper, we presented a formal model of cooperation and competition for evolutionary computation. We believe that our model constitutes the first formal, much more precise and more generic approach trying to capture the essence of cooperation and competition for evolutionary algorithms. This was possible because of precise formulation on notions of cooperation, competition, completeness, various types of optimization, an extension of the notion of decidability – all of them used in the context of several extensions of Evolutionary Turing Machine models.

We justified in our paper that evolutionary algorithms form a special case of super-recursive algorithms, and are more powerful than conventional recursive algorithms. In a similar way, Evolutionary Turing Machines are more expressive than conventional Turing Machines and belong to the family of superrecursive models of computation (Burgin, 2005, Eberbach, Goldin and Wegner, 2004). We proposed several types of evolutionary computation, including a bounded finite class, an unbounded finite class, and the most powerful – an infinite evolutionary computation class. For those types, we presented basic results on their completeness, optimality, search optimality, total optimality and decidability. We extended our model to parallel and weighted Evolutionary Turing Machines to capture properly optimization of the population trying to achieve a common goal. We demonstrated that such extension is simple and natural in our model, and allows us to capture both cooperation and competition of the whole population. As the surprising result, we obtained that cooperation fulfills a similar function to elitism to maintain optimum and speeds up convergence rate for the case of cooperating agents.

Of course, much more research is needed. It seems that it is possible and desirable to generalize our results beyond evolutionary computation search. For example, the kΩ-optimization meta-search algorithm from the $-Calculus process algebra model (Eberbach, 2005b) allows to simulate many other search algorithms (including evolutionary algorithms) and it could be used to generalize our results. This has been left for the future research.